\documentclass[conference]{IEEEtran}
\IEEEoverridecommandlockouts
\usepackage{cite}
\usepackage{amsmath,amssymb,amsfonts}
\usepackage{algorithmic}
\usepackage{graphicx}
\usepackage{textcomp}
\usepackage{xcolor}
\usepackage{xspace}
\usepackage{tikz}
\usepackage{amsthm}

\usepackage{array}
\usepackage{caption}
\usepackage{cuted}
\usepackage{float}
\usetikzlibrary{arrows.meta,positioning,shapes.geometric}
\def\BibTeX{{\rm B\kern-.05em{\sc i\kern-.025em b}\kern-.08em
    T\kern-.1667em\lower.7ex\hbox{E}\kern-.125emX}}
\newcommand{\algname}{\textsc{CaliFuse}\xspace}

\begin{document}

\title{When Foundation Models Disagree: Representational Heterogeneity for Calibrated Uncertainty in Whole-Slide Image Classification

}

\author{

\IEEEauthorblockN{
Yekai Shen\textsuperscript{2,*},
Yuqian Wang\textsuperscript{2,*},
Zishun Liao\textsuperscript{2},
Rongsheng Wang\textsuperscript{2},\\
Wenting Chen\textsuperscript{3},
S. Kevin Zhou\textsuperscript{2,5,6,\dag},
Qingsong Yao\textsuperscript{1,4,\dag}
}
\IEEEauthorblockA{
\textsuperscript{1}Renmin University of China\quad
\textsuperscript{2}University of Science and Technology of China\quad
\textsuperscript{3}Stanford University\\
\textsuperscript{4}Beijing Key Laboratory of Smart Healthcare and AI Driven Health Governance\\
\textsuperscript{5}Medical Imaging, Robotics, Analytic Computing \& Learning (MIRACLE) Lab\\
\textsuperscript{6}Jiangsu Key Laboratory of Multimodal Digital Twin Technology\\
\textsuperscript{*}These authors contributed equally.\quad
\textsuperscript{\dag}Corresponding authors: Qingsong Yao and S. Kevin Zhou.
}
}
\IEEEaftertitletext{\vspace{-5mm}}
\maketitle

\begin{abstract}
Pathology foundation models supply strong representations for whole-slide image classification, yet the confidence attached to their predictions remains hard to trust. Prevailing uncertainty estimators act within a single feature space and thereby overlook the uncertainty that the choice of foundation-model representation itself induces. This work introduces \algname{}, a calibration-oriented framework that fuses several pathology foundation models for the specific purpose of uncertainty estimation. Departing from accuracy-driven fusion, \algname{} reads representational heterogeneity across encoders as an explicit confidence cue. Slide-level decision consensus is coupled with chunk-level evidence agreement quantified through Jensen--Shannon divergence, so that both prediction stability and fine-grained evidence consistency inform the estimate. Foundation-model encoders stay frozen. A task-specific nnMIL classifier is still trained for each FM; \algname{} then estimates confidence at inference without training an extra fusion module. Evaluated on seven WSI classification tasks under several foundation-model combinations, the framework sharpens calibration, misclassification detection, and selective prediction relative to both the strongest single-encoder baseline and naive ensemble averaging, all while preserving competitive balanced accuracy.
\end{abstract}


\section{Introduction}
\begin{figure*}[t]

    \centering

    \includegraphics[width=\textwidth]{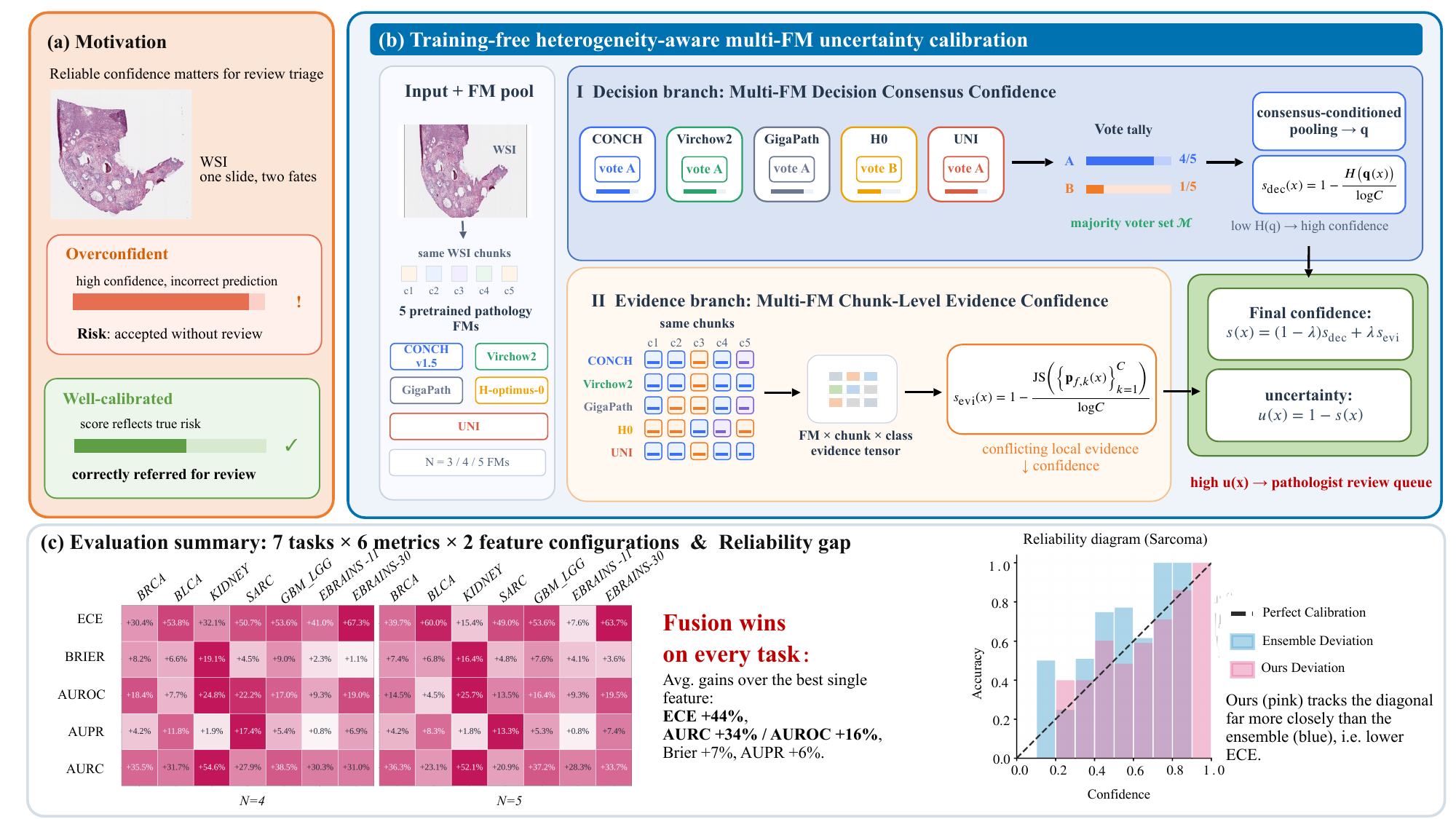}

    \caption{Overview of \algname{}.
(a) Motivation: calibrated confidence supports review triage, steering
high-uncertainty slides $u(x)$ toward pathologists rather than accepting
overconfident errors. (b) The training-free fusion stage fuses $N{=}3/4/5$
pathology FMs via a \emph{decision} branch (consensus-conditioned pooling)
and an \emph{evidence} branch (chunk-level agreement).
(c) Over seven tasks and six metrics, \algname{} improves calibration and
error ranking over the best single feature; the Sarcoma reliability diagram
shows \algname{} (pink) hugging the perfect-calibration diagonal far more
tightly than the naive ensemble (blue).}

    \label{fig:overall_pipeline}
    \vspace{-5mm}

\end{figure*}
Whole-slide image (WSI) analysis has moved away from training
slide classifiers from scratch. Contemporary pipelines encode gigapixel
tissue slides into transferable representations through pathology foundation
models (FMs), then couple these features with multiple-instance learning
(MIL) to produce slide-level predictions~\cite{ilse2018abmil,luo2025nnmil,lu2024conch,xu2024gigapath,saillard2024hoptimus,chen2024uni,vorontsov2023virchow}. Such designs
have advanced tumor subtyping, grading, and biomarker prediction. Clinical
deployment, however, is not served by accuracy in isolation. A useful model
must also recognize when its own answer is likely wrong, since the reported
confidence has to be trustworthy enough to underpin calibrated probability
estimation, error detection, and selective referral of difficult cases to
pathologists~\cite{guo2017calibration,geifman2017selective,dolezal2022uncertainty,olsson2022conformal,zhao2025rpmil}. Within
computational pathology, uncertainty consequently ceases to be an auxiliary
statistic and becomes an integral part of the prediction.

Most uncertainty estimation methods for WSI classification
remain confined to a single view. A fixed feature space is assumed, and
uncertainty is read off by perturbing the downstream predictor, resampling
patches, applying MC dropout, or invoking conformal and probabilistic MIL
variants~\cite{gal2016dropout,dolezal2022uncertainty,mehrtens2023benchmarking,linmans2023predictive,olsson2022conformal,wieslander2021conformal,zhao2025rpmil,cui2023bayesmil}. Valuable
as these strategies are, they mainly quantify instability inside one
representation. Should the selected FM encode a biased, incomplete, or overly
smooth picture of the slide, the downstream predictor can remain internally
consistent yet overconfident. Put differently, single-FM uncertainty captures
ambiguity residing within a representation while overlooking the uncertainty
that the representation choice itself introduces.

This gap is consequential because modern pathology FMs are
demonstrably heterogeneous, diverging in pretraining corpora, supervision
signals, magnification regimes, architecture, and morphological
priors~\cite{lu2024conch,xu2024gigapath,saillard2024hoptimus,chen2024uni,vorontsov2023virchow,zimmermann2024virchow2,sikaroudi2025comparing}. One and
the same WSI may therefore appear straightforward in one feature space and
ambiguous in another. Prior multi-FM efforts have largely exploited such
heterogeneity in the service of accuracy, most often by fusing or averaging
predictions across models~\cite{yang2025fusecpath,ding2025multimodal}. Averaging of this
kind, however, folds disagreement into a single posterior and can conceal
precisely the cases in which the system ought to hesitate. When strong FMs
part ways, their divergence is rarely mere noise to be damped out; more often
it constitutes a clinically meaningful indication that the case is borderline,
atypical, or underrepresented.

A different question thus arises: can disagreement among
pathology foundation models be transformed into a calibrated uncertainty
signal? The answer pursued here is affirmative. This paper introduces
\algname{}, a calibration-oriented multi-foundation-model fusion framework
that treats such disagreement as informative in its own right and erects a
calibrated confidence estimate directly upon it. Two complementary signals are
combined. At the decision level, consensus among FM-specific
nnMIL~\cite{luo2025nnmil} predictors is measured and
consensus-conditioned pooling is applied. At the evidence level,
Jensen--Shannon (JS) divergence is computed over the chunk-level predictive
distributions emitted by every FM. The former signal records whether distinct
models reach the same slide-level decision; the latter records whether they
rest on compatible class evidence across the slide even when their top-1
labels coincide.

Evaluation covers seven WSI classification tasks that span
tumor subtyping and biomarker-derived prediction, employs five heterogeneous
pathology FMs, and considers three fusion settings ($N=3,4,5$). Under the principal $N=5$
setting, \algname{} secures the best AUROC, AUPR, and AURC on all seven tasks
and the best ECE on six of the seven. Aggregated over the $N=3,4,5$ configurations, \algname{} records the
best AURC in every one of the 21 task-setting pairs, leaving balanced accuracy
competitive and frequently on par with the strongest multi-FM baselines.

The results also delineate a nuanced boundary for the method.
Its advantages are most pronounced for error ranking and selective prediction,
whereas absolute calibration retains some dataset dependence; naive averaging,
for instance, edges ahead on Kidney ECE in the principal $N=5$ setting. Rather
than a contradiction, this stands as an instructive takeaway: representational
disagreement is a potent uncertainty signal, even if deployments demanding
literal probabilities may still profit from an added post-hoc calibration
step~\cite{guo2017calibration}. Far from undermining the
argument, the observation reinforces the case for modeling heterogeneity
explicitly.

Our contributions are summarized as follows:
\begin{enumerate}
  \item Representational heterogeneity across pathology
  foundation models is identified as an explicit and hitherto underused source
  of uncertainty for WSI classification, moving multi-FM systems from
  accuracy-oriented fusion toward calibration-oriented uncertainty estimation.

  \item \algname{} operationalizes this distinction through two
  training-free confidence measures: consensus-conditioned pooling at the
  decision level and Jensen--Shannon divergence over chunk-level predictive
  distributions at the evidence level.

  \item Extensive experiments over seven WSI tasks and several
  fusion settings demonstrate that retaining cross-FM disagreement delivers
  consistently stronger calibration and error ranking than both oracle
  single-FM selection and naive probability averaging, without sacrificing
  competitive slide-level classification.

\end{enumerate}
\vspace{-1.5mm}
\section{Related Work}
\vspace{-2mm}

\subsection{Pathology Foundation Models}
\vspace{-1.5mm}
Self-supervised pretraining on large histology corpora has
established frozen patch encoders as a standard front end for WSI analysis.
UNI, trained by self-distillation on more than one hundred million patches
drawn from over one hundred thousand slides, transfers across a broad spectrum
of diagnostic tasks~\cite{chen2024uni}. Virchow and its
mixed-magnification successor Virchow2 scale vision transformers to millions of
slides with an explicit focus on scale-heterogeneous
morphology~\cite{vorontsov2023virchow,zimmermann2024virchow2}. Whole-slide
context is brought in by GigaPath, whose pretraining spans more than one
billion patches under a slide-level aggregator~\cite{xu2024gigapath}. H-optimus-0
supplies an independently pretrained self-supervised view at large
scale~\cite{saillard2024hoptimus}, while CONCH aligns
histology with language via vision-language pretraining~\cite{lu2024conch}. Although
these models cut downstream training cost and frequently transfer well, they
are far from interchangeable: disparities in pretraining data, architecture,
objective, magnification, and output space produce distinct morphological
priors and distinct failure modes. Representational-similarity analyses
corroborate that the encoders occupy measurably separate feature
spaces~\cite{sikaroudi2025comparing}. The best-performing
foundation model accordingly shifts with the dataset and the evaluation target,
which renders it natural to regard several models as complementary views of one
slide.

\subsection{Multiple Instance Learning for WSI Classification}
\vspace{-1.5mm}
WSI classification is routinely posed as multiple-instance
learning, in which a slide forms a bag of patch embeddings and only the slide
label is observed. Attention-based MIL introduced a simple yet effective
trainable aggregation scheme~\cite{ilse2018abmil}, and
subsequent methods such as CLAM, DSMIL, and TransMIL strengthened weakly
supervised learning and long-range aggregation~\cite{lu2021clam,li2021dsmil,shao2021transmil}. A
parallel thread renders MIL uncertainty-aware, for example through evidential
instance modeling or probabilistic bag embeddings~\cite{zhao2025rpmil,cui2023bayesmil}, yet
each such method still operates inside a single feature space. A more recent
view holds that the principal bottleneck in the foundation-model era lies in
the training and aggregation protocol rather than the encoder; nnMIL answers
this by offering a generalizable MIL scaffold equipped with randomized patch
and feature sampling, sliding-window inference, and built-in uncertainty
estimation~\cite{luo2025nnmil}. Throughout this study nnMIL
serves as a consistent training and inference backbone across all feature
spaces. Fixing the MIL pipeline guarantees that the dominant source of
variation is the foundation-model representation together with the confidence
fusion strategy rather than the aggregator.

\subsection{Foundation Model Fusion and Open Uncertainty Problem}
\vspace{-1.5mm}
Since prediction quality is not monopolized by any one
encoder, recent work fuses heterogeneous pathology representations to lift
slide classification. Classical ensembling, which combines independently
trained predictors, is a well-established route to improved
calibration~\cite{lakshminarayanan2017deepensembles}, and
pathology-specific fusion has carried the idea further. FuseCPath merges
multi-scale heterogeneous FMs via cluster-level patch re-embedding and
slide-level distillation to yield a stronger ensemble~\cite{yang2025fusecpath}, and
multimodal slide foundation models integrate several representation streams in
pursuit of accuracy~\cite{ding2025multimodal}. These efforts confirm that heterogeneity is a usable resource, yet their orientation remains primarily accuracy-driven, typically homogenizing multi-view variance into a singular representation. Separately,
uncertainty estimation for WSI classification has been examined mostly within
one representation, through MC dropout, conformal prediction, or probabilistic
MIL~\cite{gal2016dropout,dolezal2022uncertainty,mehrtens2023benchmarking,linmans2023predictive,olsson2022conformal,wieslander2021conformal,zhao2025rpmil,cui2023bayesmil}. The
two bodies of literature consequently stay largely disjoint. Once several
competitive foundation models are on hand, the open issue is not merely how to
fuse them for better prediction but also how to quantify uncertainty where
their evidence conflicts, given that naive probability pooling inherently smooths out the epistemic dissonance most diagnostic of borderline cases. Uncertainty within foundation-model
fusion thus remains open and insufficiently addressed, and it is precisely this
problem that the present paper targets

\section{Method}

\label{sec:method} 
\subsection{Problem Setup}

Let $x$ be a WSI with slide label $y \in \{1,\dots,C\}$, and let $F$
pathology foundation models each define their own feature space and nnMIL
classifier~\cite{luo2025nnmil}. Each FM has its own nnMIL classifier, trained separately for the downstream task. \algname{} is applied only afterward: it does not update the FM encoders or retrain those classifiers. ``Training-free'' refers only to this fusion and confidence stage. For a test slide, the goal is a prediction
$\hat y$ and a confidence score $s(x)\in[0,1]$. \algname{} has two branches:
a decision branch that measures cross-model stability through consensus
voting, and an evidence branch that measures agreement among chunk-level
class distributions.

\subsection{Per-Feature nnMIL Inference}

For each model $f\in\{1,\dots,F\}$, nnMIL represents the slide with $K$
chunks and emits a per-chunk class distribution $p_{f,k}\in\Delta^C$. We
also obtain one slide-level predictive distribution
$r_f\in\Delta^C$ from the corresponding nnMIL classifier. The chunk
predictions describe fine-scale evidence within each feature space, while
the slide-level distributions provide the model-level decisions used by the
consensus branch. These two outputs are the only inputs to the fusion stage.

\subsection{Multi-FM Decision Consensus Confidence}

For each model $f$, let
$\hat y_f=\arg\max_c r_f[c]$. Let $c^\star$ be the majority class and
$\mathcal{M}=\{f:\hat y_f=c^\star\}$ its voter set. We pool the model
distributions by consensus level:
\begin{equation}
\begin{aligned}
    q &=
    \begin{cases}
        r_{f^\star}, & |\mathcal{M}| = F, \\[3pt]
        \dfrac{1}{|\mathcal{S}|}\!\displaystyle\sum_{f \in \mathcal{S}} r_f, & 1 < |\mathcal{M}| < F, \\[10pt]
        \dfrac{1}{F}\!\displaystyle\sum_{f=1}^{F} r_f, & |\mathcal{M}| = 1,
    \end{cases}
    \\
    \hat y &= \arg\max_c q[c].
\end{aligned}
    \label{eq:decision_pooling}
\end{equation}
where $f^\star=\arg\max_{f\in\mathcal{M}} r_f[c^\star]$ is the most
confident model under full agreement, and
$\mathcal{S}\subseteq\mathcal{M}$ holds the two highest-probability models
on $c^\star$ under partial agreement. Full agreement keeps the strongest
agreeing view, partial agreement averages the two strongest agreeing views,
and full disagreement averages all views. Because each model contributes a
single slide-level distribution, the pooled distribution $q$ is also the
final decision distribution.

We convert decision stability into a confidence score using the normalized
entropy of the pooled consensus distribution,
\begin{equation}
    s_{\mathrm{dec}}
    = 1 - \frac{H(q)}{\log C}
\label{eq:decision_confidence}
\end{equation}
with $H(\cdot)$ the Shannon entropy. Low entropy means the consensus
distribution concentrates on one class and therefore yields high
$s_{\mathrm{dec}}$; diffuse pooled predictions yield lower decision
confidence. This design is consistent with the broader observation that agreement among independently trained predictors is informative about predictive reliability~\cite{lakshminarayanan2017deepensembles}: when FM-specific nnMIL predictors agree, their shared prediction is more likely to be trustworthy, and when they disagree, the resulting uncertainty should be reflected explicitly rather than resolved by an arbitrary tie-breaking rule.

\subsection{Chunk-Level Evidence Confidence}

Decision voting compares only slide-level labels. To capture finer
conflicts, we pool all chunk distributions across all models,
$\bar p = \tfrac{1}{FK}\sum_{f=1}^{F}\sum_{k=1}^{K} p_{f,k}$, and measure
their generalized JS divergence:
\begin{equation}
    \mathrm{JS}
    =
    H(\bar p)
    -
    \frac{1}{FK}\sum_{f=1}^{F}\sum_{k=1}^{K} H(p_{f,k}),
    \quad
    s_{\mathrm{evi}} = 1 - \frac{\mathrm{JS}}{\log C}
    \label{eq:evidence_confidence}
\end{equation}
Low divergence means the pooled chunk distributions are similar; high
divergence means they differ. Because the average runs over all model--chunk
pairs rather than matched chunks, this value mixes cross-model disagreement
on corresponding tissue with within-model differences across chunks of the
same slide. We therefore use it as evidence inconsistency, not as a pure
measure of representational disagreement.
Prior probabilistic MIL work shows
that uncertainty can arise below the slide
level~\cite{zhao2025rpmil,cui2023bayesmil}. Two models may reach the same
slide-level decision while relying on different regions of the slide as
evidence. Aggregating chunk-level distributions exposes this disagreement,
which slide-level voting cannot capture.

\subsection{Heterogeneity-Aware Confidence Fusion}

The two branches capture complementary aspects of confidence, and we
combine them convexly:
\begin{equation}
    s(x) = (1-\lambda)\,s_{\mathrm{dec}}(x) + \lambda\,s_{\mathrm{evi}}(x),
    \qquad
    u(x) = 1 - s(x)
    \label{eq:confidence_fusion}
\end{equation}
with $\lambda\in[0,1]$. We use $\lambda=0.5$ in all main experiments, an
equal-weight combination chosen as the simplest interpretable fusion of the
two signals rather than a tuned value; its sensitivity is studied in
Section~\ref{sec:lambda_ablation}. The final output is the prediction
$\hat y$ from Eq.~\eqref{eq:decision_pooling} and the confidence $s(x)$ from
Eq.~\eqref{eq:confidence_fusion}. Relative to one FM, the extra cost is the $N$ independently trained FM--nnMIL branches ($N{=}3$, $4$, or $5$). Fusion only combines their existing slide- and chunk-level distributions by consensus pooling, entropy, and generalized JS divergence; it does not re-encode patches or update parameters. Patch features are pre-extracted offline and are not rerun by \algname{}.

\section{Experiments}
\label{sec:experiments}

\subsection{Datasets and Preprocessing}
\label{sec:datasets}
We evaluate \algname{} on seven WSI classification tasks spanning two
clinically distinct problem types: tumor subtyping, which distinguishes
histological or molecular subtypes within a cancer type, and
biomarker-derived classification, where class labels are defined by
molecular or receptor-status endpoints with direct therapeutic relevance.
Counts, class numbers, and the evaluation split are in Table~\ref{tab:dataset_summary}.
The five cross-validated tasks are drawn from TCGA (KIDNEY, SARC, BLCA, BRCA, and LGG/GBM). EBRAINS-11 and EBRAINS-30 come from the EBRAINS Digital Brain Tumour Atlas.

\noindent\textbf{a) Tumor subtyping.}
Kidney separates clear-cell, papillary, and chromophobe carcinoma by histology.
Sarcoma covers soft-tissue subtypes including leiomyosarcoma, dedifferentiated
liposarcoma, and synovial sarcoma. BLCA-Papillary separates papillary from
non-papillary growth, a distinction read under microscopy and tied to prognosis.
EBRAINS-11 and EBRAINS-30 share one cohort: the former collapses 30 WHO classes
into 11 broad categories, and the latter keeps the full taxonomy.

\noindent\textbf{b) Biomarker-derived classification.}
BRCA-TNBC separates triple-negative from non-triple-negative breast cancer by
ER, PR, and HER2 status, which determines treatment eligibility. GBM/LGG assigns
slides to glioma subtypes defined by IDH mutation and 1p/19q codeletion under
the WHO 2021 scheme.

\noindent\textbf{c) Preprocessing.}
All experiments use pre-extracted HDF5 feature bags, with each file
storing per-patch feature vectors and their spatial coordinates. Five
tasks (BRCA-TNBC, BLCA-Papillary, Kidney, Sarcoma, GBM/LGG) use a
patient-level five-fold cross-validation split shared across all
single-view nnMIL runs. Slides from the same patient are assigned to the
same fold to prevent data leakage. The two EBRAINS tasks follow the
dataset's official split and are evaluated on the fixed test partition.
Each task is processed through five pathology foundation models, CONCH
v1.5, GigaPath, H-optimus-0, UNI, and Virchow2, yielding feature
dimensions of 768, 1{,}536, 1{,}536, 1{,}024, and 2{,}560 respectively.
Feature bags are truncated or padded to a fixed maximum sequence length
per task during training (1{,}369 to 2{,}219 patches depending on the
cohort). Full patch sequences are retained at validation and test time.

\begin{table}[t]
\vspace{-3mm}
\caption{WSI classification tasks evaluated in this study. WSI / Patients
reports slide count and unique patient count. Tasks are grouped by
prediction type.}
\label{tab:dataset_summary}
\centering
\scriptsize
\begin{tabular}{l p{2.2cm} c c c}
\hline
\textbf{Task} & \textbf{Prediction target} & \textbf{Cls} & \textbf{WSI / Pat.} & \textbf{Eval} \\
\hline
\multicolumn{5}{l}{\textit{Biomarker-derived classification}} \\
BRCA-TNBC & TNBC vs.\ non-TNBC & 2 & 721 / 641 & 5 folds \\
GBM/LGG & IDH / 1p19q glioma subtype & 4 & 1572 / 814 & 5 folds \\
\hline
\multicolumn{5}{l}{\textit{Tumor subtyping}} \\
BLCA-Pap. & Papillary vs.\ non-pap. & 2 & 452 / 380 & 5 folds \\
Kidney & Renal cancer subtype & 3 & 920 / 890 & 5 folds \\
Sarcoma & Sarcoma histology subtype & 6 & 597 / 251 & 5 folds \\
EBRAINS-11 & Coarse brain tumor category & 11 & 2319 / 2147 & official \\
EBRAINS-30 & Fine-grained brain tumor & 30 & 2319 / 2147 & official \\
\hline
\end{tabular}
\vspace{-3mm}
\end{table}

\subsection{Foundation Models and Experimental Protocol}
\label{sec:protocol}
We do not assume that the five feature spaces are aligned or share the
same dimensionality; nnMIL reads each exported HDF5 feature bag in its
native dimension. \algname{} then consumes the resulting slide-level probability vectors together with chunk-level class distributions. For each extractor and each task, nnMIL trains a
SimpleMIL model under five-fold cross-validation, so both training and nnMIL inference scale with $N$; each extractor keeps its own checkpoint. \algname{} adds no further trained module. The HDF5 bags it reads are offline features, counted as preprocessing rather than as fusion time. During fold $k$, only slides assigned
to fold $k$ are used as the test set, and all metrics are computed for
that fold before averaging across folds. The training wrapper uses batch
size 32, learning rate $3\times10^{-4}$, weight decay $10^{-4}$, and up to
100 epochs unless otherwise noted.

At inference time, we compute $s_{\mathrm{dec}}$ and $s_{\mathrm{evi}}$
as defined in Sec.~\ref{sec:method}, with no post-hoc calibration
applied. All experiments use random seed 42; each fold's output includes
chunk probabilities, voting probabilities, prediction CSV files, and a
metric report, summarized as the mean across folds. In Tables~\ref{tab:main_n5}--\ref{tab:main_n3}, five-fold entries are mean$\pm$sample standard deviation across folds. EBRAINS-11 and EBRAINS-30 use the official split and are single evaluations, so they are reported without $\pm$.

\subsection{Metric Design and Interpretation}
\label{sec:metric_design}
Our central claim is that \algname{}, by fusing multiple
foundation-model views, improves confidence estimation relative to any
single feature, while balanced accuracy stays competitive rather than being traded away. We report six metrics, fixed throughout the paper: ECE,
Brier score, AUROC, AUPR, AURC, and balanced accuracy (Bacc). ECE measures
calibration as the bin-wise gap between predicted confidence and
empirical accuracy. Brier score evaluates the squared error of the full
predictive distribution per sample, penalizing confident wrong
predictions. AUROC measures whether confidence ranks correct predictions
above incorrect ones; AUPR evaluates the same ranking property under
class imbalance. AURC sorts slides by confidence and accumulates risk as
more slides are accepted, so a low AURC means errors concentrate among
low-confidence slides. Bacc depends only on predicted labels, guarding
against apparent gains in these metrics that come from degrading the
classifier itself.
For every metric, we compare \algname{} against the best single extractor for that
metric (oracle, O) and against a naive ensemble (E), which averages both
the slide-level probability vectors and the per-slide confidence scores
of all five extractors and takes the arg-max class.
\subsection{Main Results}
\label{sec:main_results}
Table~\ref{tab:main_n5} reports the $N{=}5$ setting (best value per row in bold).
On EBRAINS-30, the largest multi-class task, AURC is the most informative metric since the system must rank high-risk slides lower in the review queue amid heavy class competition. Our \algname{} reduces AURC by 33.7\% and 36.8\% relative to the single-extractor oracle and naive ensemble, respectively, and also improves ECE (63.6\%), Brier (3.7\%), AUROC (19.5\%), and AUPR (7.4\%) over the oracle, confirming that cross-view disagreement is especially valuable when class competition is high. GBM/LGG further shows the gains hold under merged cohorts: relative to the oracle, ECE, Brier, AUROC, AUPR, and AURC improve by 53.5\%, 7.5\%, 16.4\%, 5.4\%, and 36.6\%, respectively. BRCA-TNBC offers a complementary binary, biomarker-driven setting, where AUPR and AUROC best reflect positive-class recognition and discrimination. Our \algname{} improves AUPR and AUROC by 4.2\% and 14.5\% over the oracle and by 7.2\% and 36.7\% over the naive ensemble, confirming the \algname{} estimate generalizes to binary biomarker classification.

\begin{table*}[htbp]
\vspace{-3mm}
\caption{$N=5$ foundation models (UNI + GigaPath + H-optimus-0 + Virchow2 + CONCH v1.5). Oracle = best single-extractor oracle for each metric; Ensemble = naive ensemble averaging of the $N$ extractors' slide-level probabilities; \algname{} = our proposed fusion method. For ECE, Brier, and AURC, lower is better; for AUROC, AUPR, and Bacc, higher is better. Bold marks the best value in each row. $^{*}$Bacc is an accuracy metric, rather than an uncertainty calibration metric.}
\label{tab:main_n5}
\centering
\resizebox{\textwidth}{!}{%
\begin{tabular}{l |ccc| ccc| ccc}
\hline
& \multicolumn{3}{c|}{\textbf{ECE}$\downarrow$} & \multicolumn{3}{c|}{\textbf{Brier}$\downarrow$} & \multicolumn{3}{c}{\textbf{AUROC}$\uparrow$} \\
\textbf{Task} & Oracle & Ensemble & \algname{} & Oracle & Ensemble & \algname{} & Oracle & Ensemble & \algname{} \\
\hline
BRCA-TNBC & 0.098$\pm$0.028 & 0.089$\pm$0.036 & \textbf{0.059$\pm$0.009} & 0.230$\pm$0.039 & 0.225$\pm$0.042 & \textbf{0.212$\pm$0.045} & 0.641$\pm$0.103 & 0.537$\pm$0.076 & \textbf{0.734$\pm$0.091} \\
BLCA-Papillary & 0.210$\pm$0.043 & 0.166$\pm$0.026 & \textbf{0.084$\pm$0.014} & 0.397$\pm$0.028 & 0.389$\pm$0.034 & \textbf{0.370$\pm$0.041} & 0.559$\pm$0.060 & 0.483$\pm$0.111 & \textbf{0.584$\pm$0.040} \\
Kidney & 0.050$\pm$0.013 & \textbf{0.034$\pm$0.011} & 0.042$\pm$0.016 & 0.111$\pm$0.027 & 0.115$\pm$0.022 & \textbf{0.093$\pm$0.015} & 0.677$\pm$0.071 & 0.641$\pm$0.090 & \textbf{0.851$\pm$0.048} \\
Sarcoma & 0.344$\pm$0.070 & 0.335$\pm$0.077 & \textbf{0.175$\pm$0.019} & 0.559$\pm$0.068 & 0.562$\pm$0.021 & \textbf{0.532$\pm$0.047} & 0.586$\pm$0.081 & 0.525$\pm$0.081 & \textbf{0.665$\pm$0.111} \\
GBM/LGG & 0.155$\pm$0.063 & 0.129$\pm$0.057 & \textbf{0.072$\pm$0.025} & 0.308$\pm$0.070 & 0.295$\pm$0.074 & \textbf{0.285$\pm$0.080} & 0.707$\pm$0.011 & 0.706$\pm$0.021 & \textbf{0.823$\pm$0.031} \\
EBRAINS-11 & 0.038 & 0.036 & \textbf{0.035} & 0.136 & 0.166 & \textbf{0.130} & 0.807 & 0.746 & \textbf{0.882} \\
EBRAINS-30 & 0.209 & 0.190 & \textbf{0.076} & 0.380 & 0.387 & \textbf{0.366} & 0.651 & 0.587 & \textbf{0.778} \\
\hline
& \multicolumn{3}{c|}{\textbf{AUPR}$\uparrow$} & \multicolumn{3}{c|}{\textbf{AURC}$\downarrow$} & \multicolumn{3}{c}{\textbf{Bacc}$^{*}\uparrow$} \\
\textbf{Task} & Oracle & Ensemble & \algname{} & Oracle & Ensemble & \algname{} & Oracle & Ensemble & \algname{} \\
\hline
BRCA-TNBC & 0.901$\pm$0.060 & 0.876$\pm$0.045 & \textbf{0.939$\pm$0.027} & 0.104$\pm$0.059 & 0.126$\pm$0.045 & \textbf{0.066$\pm$0.027} & 0.832$\pm$0.059 & \textbf{0.835$\pm$0.053} & 0.819$\pm$0.060 \\
BLCA-Papillary & 0.743$\pm$0.051 & 0.732$\pm$0.083 & \textbf{0.804$\pm$0.046} & 0.262$\pm$0.052 & 0.275$\pm$0.086 & \textbf{0.201$\pm$0.044} & \textbf{0.689$\pm$0.083} & 0.679$\pm$0.033 & 0.686$\pm$0.022 \\
Kidney & 0.968$\pm$0.012 & 0.967$\pm$0.012 & \textbf{0.986$\pm$0.007} & 0.033$\pm$0.012 & 0.034$\pm$0.012 & \textbf{0.016$\pm$0.006} & 0.954$\pm$0.011 & \textbf{0.957$\pm$0.010} & 0.956$\pm$0.010 \\
Sarcoma & 0.679$\pm$0.063 & 0.659$\pm$0.103 & \textbf{0.769$\pm$0.077} & 0.340$\pm$0.050 & 0.357$\pm$0.097 & \textbf{0.269$\pm$0.069} & 0.616$\pm$0.089 & \textbf{0.644$\pm$0.051} & 0.639$\pm$0.048 \\
GBM/LGG & 0.897$\pm$0.041 & 0.906$\pm$0.046 & \textbf{0.945$\pm$0.029} & 0.112$\pm$0.044 & 0.102$\pm$0.050 & \textbf{0.071$\pm$0.037} & 0.737$\pm$0.098 & \textbf{0.745$\pm$0.105} & 0.741$\pm$0.109 \\
EBRAINS-11 & 0.978 & 0.970 & \textbf{0.985} & 0.024 & 0.031 & \textbf{0.018} & 0.921 & \textbf{0.931} & \textbf{0.931} \\
EBRAINS-30 & 0.848 & 0.833 & \textbf{0.911} & 0.166 & 0.174 & \textbf{0.110} & 0.719 & 0.752 & \textbf{0.753} \\
\hline
\end{tabular}}
\end{table*}
\vspace{-3mm}

\subsection{What the Results Mean for the Pipeline}
\label{sec:result_interpretation}
The results separate two conclusions. First, on balanced accuracy, simply
using multiple features (naive ensemble) already outperforms the
single-extractor oracle, showing that more views help accuracy. Second,
this advantage does not carry over to uncertainty: on the five
calibration and error-ranking metrics (ECE, Brier, AUROC, AUPR, AURC), the
naive ensemble
shows no consistent gain over the oracle, and sometimes performs worse.
Our \algname{} closes this gap, clearly outperforming both the
single-extractor oracle and the naive ensemble on all five calibration and
error-ranking metrics.

\subsection{Combination Analysis}
\label{sec:combination}
We further test whether the $N{=}5$ pattern holds with fewer views
(Tables~\ref{tab:main_n4} and~\ref{tab:main_n3}).
\begin{table*}[htbp]
\vspace{-3mm}
\caption{$N=4$ foundation models (GigaPath + H-optimus-0 + Virchow2 + CONCH v1.5). Notation follows Table~\ref{tab:main_n5}.}
\label{tab:main_n4}
\centering
\resizebox{\textwidth}{!}{%
\begin{tabular}{l |ccc| ccc| ccc}
\hline
& \multicolumn{3}{c|}{\textbf{ECE}$\downarrow$} & \multicolumn{3}{c|}{\textbf{Brier}$\downarrow$} & \multicolumn{3}{c}{\textbf{AUROC}$\uparrow$} \\
\textbf{Task} & Oracle & Ensemble & \algname{} & Oracle & Ensemble & \algname{} & Oracle & Ensemble & \algname{} \\
\hline
BRCA-TNBC & 0.098$\pm$0.028 & 0.086$\pm$0.029 & \textbf{0.068$\pm$0.013} & 0.230$\pm$0.039 & 0.219$\pm$0.041 & \textbf{0.211$\pm$0.045} & 0.641$\pm$0.103 & 0.604$\pm$0.050 & \textbf{0.759$\pm$0.086} \\
BLCA-Papillary & 0.210$\pm$0.043 & 0.170$\pm$0.036 & \textbf{0.097$\pm$0.027} & 0.397$\pm$0.028 & 0.392$\pm$0.041 & \textbf{0.371$\pm$0.065} & 0.559$\pm$0.060 & 0.486$\pm$0.118 & \textbf{0.602$\pm$0.110} \\
Kidney & 0.050$\pm$0.013 & 0.040$\pm$0.010 & \textbf{0.034$\pm$0.020} & 0.111$\pm$0.027 & 0.111$\pm$0.016 & \textbf{0.090$\pm$0.016} & 0.677$\pm$0.071 & 0.648$\pm$0.089 & \textbf{0.845$\pm$0.055} \\
Sarcoma & 0.344$\pm$0.070 & 0.346$\pm$0.091 & \textbf{0.169$\pm$0.037} & 0.559$\pm$0.068 & 0.558$\pm$0.032 & \textbf{0.534$\pm$0.056} & 0.586$\pm$0.081 & 0.538$\pm$0.064 & \textbf{0.716$\pm$0.104} \\
GBM/LGG & 0.155$\pm$0.063 & 0.129$\pm$0.056 & \textbf{0.072$\pm$0.028} & 0.308$\pm$0.070 & 0.295$\pm$0.079 & \textbf{0.281$\pm$0.076} & 0.707$\pm$0.011 & 0.694$\pm$0.021 & \textbf{0.827$\pm$0.035} \\
EBRAINS-11 & 0.057 & 0.044 & \textbf{0.034} & 0.136 & 0.166 & \textbf{0.133} & 0.807 & 0.736 & \textbf{0.881} \\
EBRAINS-30 & 0.209 & 0.196 & \textbf{0.068} & 0.380 & 0.387 & \textbf{0.376} & 0.651 & 0.578 & \textbf{0.774} \\
\hline
& \multicolumn{3}{c|}{\textbf{AUPR}$\uparrow$} & \multicolumn{3}{c|}{\textbf{AURC}$\downarrow$} & \multicolumn{3}{c}{\textbf{Bacc}$^{*}\uparrow$} \\
\textbf{Task} & Oracle & Ensemble & \algname{} & Oracle & Ensemble & \algname{} & Oracle & Ensemble & \algname{} \\
\hline
BRCA-TNBC & 0.901$\pm$0.060 & 0.898$\pm$0.034 & \textbf{0.938$\pm$0.028} & 0.104$\pm$0.059 & 0.105$\pm$0.034 & \textbf{0.067$\pm$0.028} & \textbf{0.832$\pm$0.059} & 0.831$\pm$0.054 & 0.824$\pm$0.057 \\
BLCA-Papillary & 0.743$\pm$0.051 & 0.735$\pm$0.073 & \textbf{0.830$\pm$0.025} & 0.262$\pm$0.052 & 0.272$\pm$0.077 & \textbf{0.179$\pm$0.030} & 0.689$\pm$0.083 & \textbf{0.692$\pm$0.029} & 0.675$\pm$0.041 \\
Kidney & 0.968$\pm$0.012 & 0.969$\pm$0.009 & \textbf{0.986$\pm$0.006} & 0.033$\pm$0.012 & 0.032$\pm$0.009 & \textbf{0.015$\pm$0.006} & 0.954$\pm$0.011 & 0.959$\pm$0.008 & \textbf{0.960$\pm$0.007} \\
Sarcoma & 0.679$\pm$0.063 & 0.665$\pm$0.114 & \textbf{0.797$\pm$0.082} & 0.340$\pm$0.050 & 0.351$\pm$0.110 & \textbf{0.245$\pm$0.077} & 0.616$\pm$0.089 & \textbf{0.633$\pm$0.052} & 0.622$\pm$0.051 \\
GBM/LGG & 0.897$\pm$0.041 & 0.902$\pm$0.043 & \textbf{0.945$\pm$0.031} & 0.112$\pm$0.044 & 0.106$\pm$0.047 & \textbf{0.069$\pm$0.037} & 0.737$\pm$0.098 & \textbf{0.744$\pm$0.099} & 0.740$\pm$0.106 \\
EBRAINS-11 & 0.978 & 0.969 & \textbf{0.986} & 0.024 & 0.033 & \textbf{0.017} & 0.921 & \textbf{0.931} & 0.928 \\
EBRAINS-30 & 0.848 & 0.825 & \textbf{0.907} & 0.166 & 0.182 & \textbf{0.115} & 0.715 & 0.747 & \textbf{0.748} \\
\hline
\end{tabular}}
\vspace{-3mm}
\end{table*}

\begin{table*}[htbp]

\caption{$N=3$ foundation models (GigaPath + Virchow2 + CONCH v1.5). Notation follows Table~\ref{tab:main_n5}.}
\label{tab:main_n3}
\centering
\resizebox{\textwidth}{!}{%
\begin{tabular}{l |ccc| ccc| ccc}
\hline
& \multicolumn{3}{c|}{\textbf{ECE}$\downarrow$} & \multicolumn{3}{c|}{\textbf{Brier}$\downarrow$} & \multicolumn{3}{c}{\textbf{AUROC}$\uparrow$} \\
\textbf{Task} & Oracle & Ensemble & \algname{} & Oracle & Ensemble & \algname{} & Oracle & Ensemble & \algname{} \\
\hline
BRCA-TNBC & 0.098$\pm$0.028 & 0.091$\pm$0.025 & \textbf{0.049$\pm$0.016} & 0.230$\pm$0.039 & 0.223$\pm$0.033 & \textbf{0.212$\pm$0.045} & 0.641$\pm$0.103 & 0.609$\pm$0.024 & \textbf{0.737$\pm$0.073} \\
BLCA-Papillary & 0.210$\pm$0.043 & 0.161$\pm$0.058 & \textbf{0.106$\pm$0.022} & 0.401$\pm$0.046 & 0.396$\pm$0.047 & \textbf{0.378$\pm$0.065} & 0.559$\pm$0.060 & 0.525$\pm$0.087 & \textbf{0.572$\pm$0.044} \\
Kidney & 0.050$\pm$0.013 & 0.040$\pm$0.012 & \textbf{0.035$\pm$0.013} & 0.111$\pm$0.027 & 0.113$\pm$0.022 & \textbf{0.093$\pm$0.021} & 0.677$\pm$0.071 & 0.637$\pm$0.067 & \textbf{0.852$\pm$0.063} \\
Sarcoma & 0.375$\pm$0.102 & 0.349$\pm$0.088 & \textbf{0.169$\pm$0.039} & 0.559$\pm$0.068 & 0.564$\pm$0.038 & \textbf{0.540$\pm$0.055} & 0.586$\pm$0.081 & 0.537$\pm$0.052 & \textbf{0.743$\pm$0.075} \\
GBM/LGG & 0.159$\pm$0.056 & 0.131$\pm$0.060 & \textbf{0.064$\pm$0.022} & 0.308$\pm$0.070 & 0.297$\pm$0.076 & \textbf{0.284$\pm$0.071} & 0.707$\pm$0.011 & 0.680$\pm$0.049 & \textbf{0.822$\pm$0.043} \\
EBRAINS-11 & 0.057 & 0.046 & \textbf{0.026} & 0.136 & 0.160 & \textbf{0.130} & 0.807 & 0.710 & \textbf{0.874} \\
EBRAINS-30 & 0.209 & 0.196 & \textbf{0.072} & 0.380 & 0.373 & \textbf{0.366} & 0.651 & 0.604 & \textbf{0.758} \\
\hline
& \multicolumn{3}{c|}{\textbf{AUPR}$\uparrow$} & \multicolumn{3}{c|}{\textbf{AURC}$\downarrow$} & \multicolumn{3}{c}{\textbf{Bacc}$^{*}\uparrow$} \\
\textbf{Task} & Oracle & Ensemble & \algname{} & Oracle & Ensemble & \algname{} & Oracle & Ensemble & \algname{} \\
\hline
BRCA-TNBC & 0.901$\pm$0.060 & 0.904$\pm$0.015 & \textbf{0.929$\pm$0.027} & 0.104$\pm$0.059 & 0.099$\pm$0.016 & \textbf{0.076$\pm$0.027} & 0.822$\pm$0.051 & \textbf{0.829$\pm$0.055} & 0.820$\pm$0.053 \\
BLCA-Papillary & 0.743$\pm$0.051 & 0.752$\pm$0.076 & \textbf{0.792$\pm$0.071} & 0.262$\pm$0.052 & 0.253$\pm$0.078 & \textbf{0.212$\pm$0.070} & 0.689$\pm$0.083 & 0.690$\pm$0.039 & \textbf{0.706$\pm$0.032} \\
Kidney & 0.968$\pm$0.012 & 0.969$\pm$0.010 & \textbf{0.985$\pm$0.010} & 0.033$\pm$0.012 & 0.032$\pm$0.010 & \textbf{0.016$\pm$0.010} & 0.954$\pm$0.011 & \textbf{0.961$\pm$0.010} & 0.958$\pm$0.009 \\
Sarcoma & 0.675$\pm$0.126 & 0.667$\pm$0.107 & \textbf{0.806$\pm$0.072} & 0.344$\pm$0.121 & 0.347$\pm$0.107 & \textbf{0.237$\pm$0.078} & 0.616$\pm$0.089 & 0.626$\pm$0.075 & \textbf{0.635$\pm$0.080} \\
GBM/LGG & 0.897$\pm$0.041 & 0.899$\pm$0.047 & \textbf{0.946$\pm$0.030} & 0.112$\pm$0.044 & 0.108$\pm$0.050 & \textbf{0.069$\pm$0.037} & 0.737$\pm$0.098 & \textbf{0.759$\pm$0.112} & 0.758$\pm$0.112 \\
EBRAINS-11 & 0.978 & 0.968 & \textbf{0.986} & 0.024 & 0.033 & \textbf{0.017} & 0.921 & \textbf{0.935} & 0.934 \\
EBRAINS-30 & 0.848 & 0.844 & \textbf{0.906} & 0.166 & 0.164 & \textbf{0.113} & 0.714 & \textbf{0.745} & 0.743 \\
\hline
\end{tabular}}
\vspace{-4mm}
\end{table*}

Across $N{=}3$, $N{=}4$, and $N{=}5$, \algname{} keeps its advantage over the single-extractor oracle on all seven tasks, and the advantage over the naive ensemble holds for nearly every task and $N$. On EBRAINS-30 with only three foundation models, \algname{} improves ECE by 65.6\% and 63.3\% relative to the oracle and the ensemble, AURC by 31.9\% and 31.1\%, and AUROC by 16.4\% and 25.5\%, so the gains do not require all five FMs.

\subsection{Ablation: Sensitivity to $\lambda$}
\label{sec:lambda_ablation}
\begin{figure}[H]
\centering
\includegraphics[width=\columnwidth]{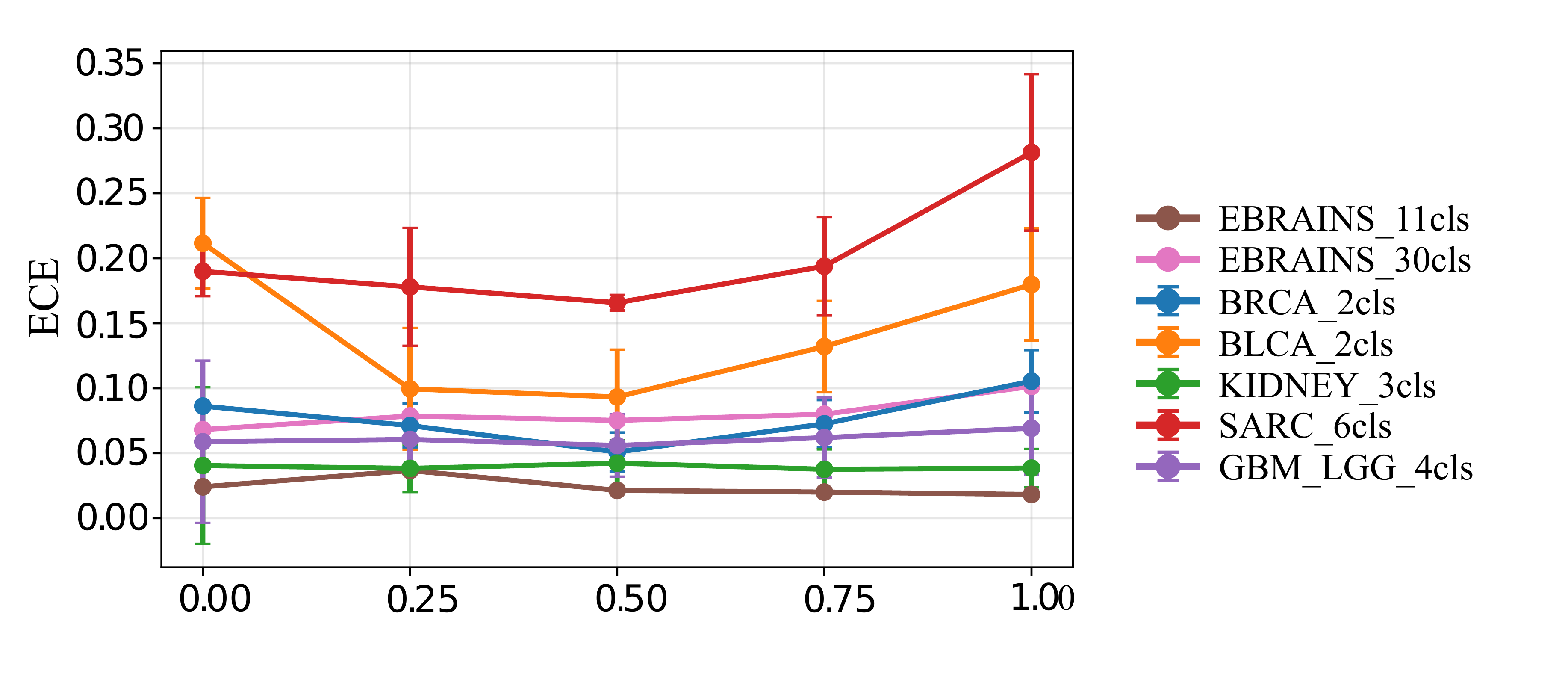}
\vspace{-5mm}
\caption{ECE as a function of the fusion weight $\lambda$, swept from 0 to
1 across all seven tasks. Error bars denote standard deviation across
folds. $\lambda=0$ corresponds to using only the decision-level signal
$s_{\mathrm{dec}}$; $\lambda=1$ corresponds to using only the
evidence-level signal $s_{\mathrm{evi}}$.}
\label{fig:lambda_sweep}
\vspace{-3mm}
\end{figure}

Figure~\ref{fig:lambda_sweep} sweeps $\lambda$ from 0 to 1 and reports ECE on all seven tasks. ECE generally improves toward the middle of the range and degrades at both extremes ($\lambda \in \{0,1\}$), confirming that both signals contribute.
\vspace{-5mm}

\subsection{Visual and Component Analysis}
\label{sec:visual_analysis}
Figure~\ref{fig:reliability} shows reliability diagrams for GBM/LGG, Sarcoma, and EBRAINS-11. The naive ensemble's bars sit above the perfect-calibration diagonal in nearly every confidence bin, so predicted confidence underestimates accuracy, whereas \algname{} tracks the diagonal more closely. Averaging tends to understate confidence when the models disagree, and folding that disagreement into $s(x)$ shifts those bins back toward the diagonal.

\begin{strip}
\centering
\includegraphics[width=0.8\textwidth]{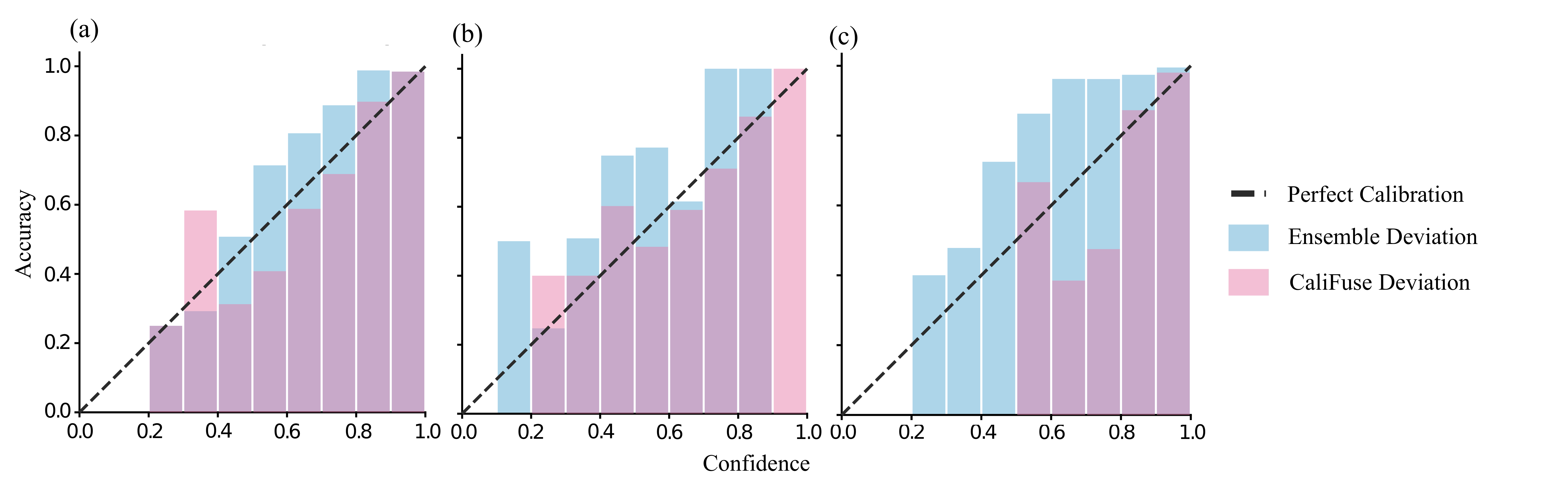}
\captionof{figure}{Reliability diagrams for (a) GBM/LGG, (b) Sarcoma, and (c)
EBRAINS-11, comparing Ensemble (blue) and \algname{} (pink/purple overlay)
against the dashed diagonal of perfect calibration.}
\label{fig:reliability}
\end{strip}

\section{Discussion and Limitations}
\label{sec:discussion}

  The findings imply that heterogeneous foundation models are
  valuable beyond the classification features they contribute: they also expose
  multi-view disagreement from which confidence can be built. A lone extractor
  may be accurate yet brittle, assigning a high softmax score even where
  differing foundation-model views or chunk-level evidence fail to endorse the
  same decision. The \algname{} fusion score is engineered to systematically penalize multi-view epistemic mismatch, which explains why the
  sharpest gains land in AUROC and AURC, the metrics that directly test whether
  confidence can push likely errors lower in the review queue.

Limitations: the gains require the $N$ trained FM--nnMIL branches, so cost grows with $N$; each $N$ is evaluated on a single FM subset; EBRAINS-11 and EBRAINS-30 come from one official split and are reported without fold variance; and $\lambda{=}0.5$ is fixed rather than tuned per task. A natural extension appends a validation-set calibration layer after fusion, such as temperature scaling~\cite{guo2017calibration}, when a deployment needs the scores to read as calibrated probabilities of correctness.

\section{Conclusion}
\label{sec:conclusion}

This paper has established that heterogeneity across pathology
foundation models, long dismissed as a nuisance to be fused away, can instead be
measured and turned into calibrated confidence. \algname{} accomplishes this
through two training-free measures operating at separate levels, Multi-FM
Decision Consensus Confidence and Multi-FM Chunk-Level Evidence Confidence,
which jointly capture slide-level decision agreement and chunk-level evidence
consistency. Experiments over seven WSI tasks confirm that representational
disagreement serves as a useful signal for calibrated confidence estimation,
advancing calibration, error ranking, and selective prediction beyond single-FM
and naive ensemble baselines while retaining competitive classification
accuracy.

\bibliographystyle{IEEEtran}
\bibliography{references}

@inproceedings{ilse2018abmil,
  author    = {Ilse, Maximilian and Tomczak, Jakub M. and Welling, Max},
  title     = {Attention-based deep multiple instance learning},
  booktitle = {Proc. Int. Conf. Mach. Learn. (ICML)},
  year      = {2018}
}

@article{luo2025nnmil,
  author  = {Luo, Xiangde and Xiang, Jinxi and Ji, Yuanfeng and Li, Ruijiang},
  title   = {{nnMIL}: A generalizable multiple instance learning framework for computational pathology},
  journal = {arXiv preprint arXiv:2511.14907},
  year    = {2025}
}

@article{lu2024conch,
  author  = {Lu, Ming Y. and Chen, Bowen and Williamson, Drew F. K. and Chen, Richard J. and Liang, Ivy and Ding, Tong and Jaume, Guillaume and Parwani, Anil V. and Zhang, Andrew and Mahmood, Faisal},
  title   = {A visual-language foundation model for computational pathology},
  journal = {Nat. Med.},
  volume  = {30},
  pages   = {863--874},
  year    = {2024}
}

@article{xu2024gigapath,
  author  = {Xu, Hanwen and Usuyama, Naoto and Bagga, Jaspreet and Zhang, Sheng and Rao, Rajesh and Naumann, Tristan and Wong, Cliff and Gero, Zelalem and Gonzalez, Javier and Gu, Yu and Poon, Hoifung},
  title   = {A whole-slide foundation model for digital pathology from real-world data},
  journal = {Nature},
  volume  = {630},
  pages   = {181--188},
  year    = {2024}
}

@misc{saillard2024hoptimus,
  author       = {Saillard, Charlie and Jenatton, Rodolphe and Llinares-Lopez, Felipe and Mariet, Zelda and Cahane, David and Durand, Eric and Vert, Jean-Philippe},
  title        = {{H-optimus-0}},
  year         = {2024},
  howpublished = {\url{https://github.com/bioptimus/releases/tree/main/models/h-optimus/v0}}
}

@article{chen2024uni,
  author  = {Chen, Richard J. and Ding, Tong and Lu, Ming Y. and Williamson, Drew F. K. and Jaume, Guillaume and Chen, Bowen and Zhang, Andrew and Shao, Daniel and Song, Andrew H. and Mahmood, Faisal},
  title   = {Towards a general-purpose foundation model for computational pathology},
  journal = {Nat. Med.},
  volume  = {30},
  pages   = {850--862},
  year    = {2024}
}

@article{vorontsov2023virchow,
  author  = {Vorontsov, Eugene and Casson, Adam and Zimmermann, Eric and others},
  title   = {Virchow: A million-slide digital pathology foundation model},
  journal = {arXiv preprint arXiv:2309.07778},
  year    = {2023}
}

@inproceedings{guo2017calibration,
  author    = {Guo, Chuan and Pleiss, Geoff and Sun, Yu and Weinberger, Kilian Q.},
  title     = {On calibration of modern neural networks},
  booktitle = {Proc. Int. Conf. Mach. Learn. (ICML)},
  year      = {2017}
}

@inproceedings{gal2016dropout,
  author    = {Gal, Yarin and Ghahramani, Zoubin},
  title     = {Dropout as a {Bayesian} approximation: Representing model uncertainty in deep learning},
  booktitle = {Proc. Int. Conf. Mach. Learn. (ICML)},
  year      = {2016}
}

@inproceedings{geifman2017selective,
  author    = {Geifman, Yonatan and El-Yaniv, Ran},
  title     = {Selective classification for deep neural networks},
  booktitle = {Proc. Adv. Neural Inf. Process. Syst. (NeurIPS)},
  year      = {2017}
}

@inproceedings{lakshminarayanan2017deepensembles,
  author    = {Lakshminarayanan, Balaji and Pritzel, Alexander and Blundell, Charles},
  title     = {Simple and scalable predictive uncertainty estimation using deep ensembles},
  booktitle = {Proc. Adv. Neural Inf. Process. Syst. (NeurIPS)},
  year      = {2017}
}

@article{lu2021clam,
  author  = {Lu, Ming Y. and Williamson, Drew F. K. and Chen, Tiffany Y. and Chen, Richard J. and Barbieri, Matteo and Mahmood, Faisal},
  title   = {Data-efficient and weakly supervised computational pathology on whole-slide images},
  journal = {Nat. Biomed. Eng.},
  volume  = {5},
  pages   = {555--570},
  year    = {2021}
}

@inproceedings{li2021dsmil,
  author    = {Li, Bin and Li, Yin and Eliceiri, Kevin W.},
  title     = {Dual-stream multiple instance learning network for whole slide image classification with self-supervised contrastive learning},
  booktitle = {Proc. IEEE/CVF Conf. Comput. Vis. Pattern Recognit. (CVPR)},
  pages     = {14318--14328},
  year      = {2021}
}

@inproceedings{shao2021transmil,
  author    = {Shao, Zhuchen and Bian, Hao and Chen, Yang and Wang, Yifeng and Zhang, Jian and Ji, Xiangyang and Zhang, Yongbing},
  title     = {{TransMIL}: Transformer based correlated multiple instance learning for whole slide image classification},
  booktitle = {Proc. Adv. Neural Inf. Process. Syst. (NeurIPS)},
  year      = {2021}
}

@article{dolezal2022uncertainty,
  author  = {Dolezal, James M. and Srisuwananukorn, Andrew and Karpeyev, Dmitry and Kleinsmith, M. A. R.},
  title   = {Uncertainty-informed deep learning models enable high-confidence predictions for digital histopathology},
  journal = {Nat. Commun.},
  volume  = {13},
  pages   = {6572},
  year    = {2022}
}

@article{mehrtens2023benchmarking,
  author  = {Mehrtens, Hendrik A. and Kurz, Alexander and Bucher, Tabea-Clara and Brinker, Titus J.},
  title   = {Benchmarking common uncertainty estimation methods with histopathological images under domain shift and label noise},
  journal = {Med. Image Anal.},
  volume  = {89},
  pages   = {102914},
  year    = {2023}
}

@article{linmans2023predictive,
  author  = {Linmans, Jasper and Elfwing, Stefan and van der Laak, Jeroen and Litjens, Geert},
  title   = {Predictive uncertainty estimation for out-of-distribution detection in digital pathology},
  journal = {Med. Image Anal.},
  volume  = {83},
  pages   = {102655},
  year    = {2023}
}

@article{olsson2022conformal,
  author  = {Olsson, Henrik and Kartasalo, Kimmo and Mulliqi, Nita and others},
  title   = {Estimating diagnostic uncertainty in artificial intelligence assisted pathology using conformal prediction},
  journal = {Nat. Commun.},
  volume  = {13},
  pages   = {7761},
  year    = {2022}
}

@article{wieslander2021conformal,
  author  = {Wieslander, H{\aa}kan and Harrison, Philip J. and Skogberg, Gabriel and others},
  title   = {Deep learning with conformal prediction for hierarchical analysis of large-scale whole-slide tissue images},
  journal = {IEEE J. Biomed. Health Inform.},
  volume  = {25},
  number  = {2},
  pages   = {371--380},
  year    = {2021}
}

@inproceedings{zhao2025rpmil,
  author    = {Zhao, Zhihua and Chen, Kun and Zhao, Jian},
  title     = {{RPMIL}: Rethinking uncertainty-aware probabilistic multiple instance learning for whole slide pathology diagnosis},
  booktitle = {Proc. Int. Joint Conf. Artif. Intell. (IJCAI)},
  pages     = {2467--2475},
  year      = {2025}
}

@inproceedings{cui2023bayesmil,
  author    = {Cui, Yufei and Liu, Ziquan and Liu, Xiangyu and Liu, Xue and Wang, Cong and Kuo, Tei-Wei and Xue, Chun Jason and Chan, Antoni B.},
  title     = {{Bayes-MIL}: A new probabilistic perspective on attention-based multiple instance learning for whole slide images},
  booktitle = {Proc. Int. Conf. Learn. Represent. (ICLR)},
  year      = {2023}
}

@article{ding2025multimodal,
  author  = {Ding, Tong and Wagner, Sophia J. and Song, Andrew H. and Chen, Richard J. and Lu, Ming Y. and Zhang, Andrew and Vaidya, Anurag J. and Jaume, Guillaume and Shaban, Muhammad and Kim, Ahrong and Williamson, Drew F. K. and Chen, Bowen and Almagro-Perez, Cristina and Doucet, Paul and Sahai, Sharifa and Chen, Chengkuan and Komura, Daisuke and Kawabe, Akihiro and Ishikawa, Shumpei and Gerber, Georg and Peng, Tingying and Le, Long Phi and Mahmood, Faisal},
  title   = {Multimodal whole slide foundation model for pathology},
  journal = {Nat. Med.},
  year    = {2025}
}

@article{zimmermann2024virchow2,
  author  = {Zimmermann, Eric and Vorontsov, Eugene and Viret, Julian and Casson, Adam and Zelechowski, Michal and Shaikovski, George and Tenenholtz, Neil and Hall, James and Fuchs, Thomas and Fusi, Nicolo and Liu, Siqi and Severson, Kristen},
  title   = {{Virchow2}: Scaling self-supervised mixed magnification models in pathology},
  journal = {arXiv preprint arXiv:2408.00738},
  year    = {2024}
}

@article{sikaroudi2025comparing,
  author  = {Sikaroudi, Amir and others},
  title   = {Comparing computational pathology foundation models using representational similarity analysis},
  journal = {arXiv preprint arXiv:2509.15482},
  year    = {2025}
}

@article{yang2025fusecpath,
  author  = {Yang, Zhidong and Shi, Xiaoqing and Ba, Wei and Song, Zheng and Luan, Hao and Hu, Tian and Lin, Shuo and Wang, Jie and Zhou, S. Kevin and Yan, Rui},
  title   = {Fusion of multi-scale heterogeneous pathology foundation models for whole slide image analysis},
  journal = {arXiv preprint arXiv:2510.27237},
  year    = {2025}
}
\end{document}